\documentclass[10pt,twocolumn,letterpaper]{article}

\usepackage[pagenumbers]{cvpr} 

\usepackage{soul}
\usepackage{utfsym}
\usepackage{multirow}
\definecolor{darkgreen}{rgb}{0.0, 0.5, 0.0}
\definecolor{darkred}{rgb}{0.6, 0.0, 0.0}
\newcommand{\newcheckmark}{\textcolor{darkgreen}{\usym{2713}}}
\newcommand{\newcrossmark}{\textcolor{darkred}{\usym{2717}}}
\definecolor{cvprblue}{rgb}{0.21,0.49,0.74}
\usepackage[pagebackref,breaklinks,colorlinks,allcolors=cvprblue]{hyperref}

\def\paperID{***} 
\def\confName{CVPR}
\def\confYear{2025}

\title{Instruction-based Image Manipulation by Watching How Things Move}

\author{
Mingdeng~Cao$^{1}$\quad
Xuaner~Zhang$^2$\quad
Yinqiang~Zheng$^1$\quad
Zhihao~Xia$^2$\\
\\
$^1$The University of Tokyo\quad
$^2$Adobe \\ 
}

\begin{document}
\twocolumn[{%
\renewcommand\twocolumn[1][]{#1}%
\maketitle
\begin{center}
    \centering
    \captionsetup{type=figure}
    \includegraphics[width=\linewidth]{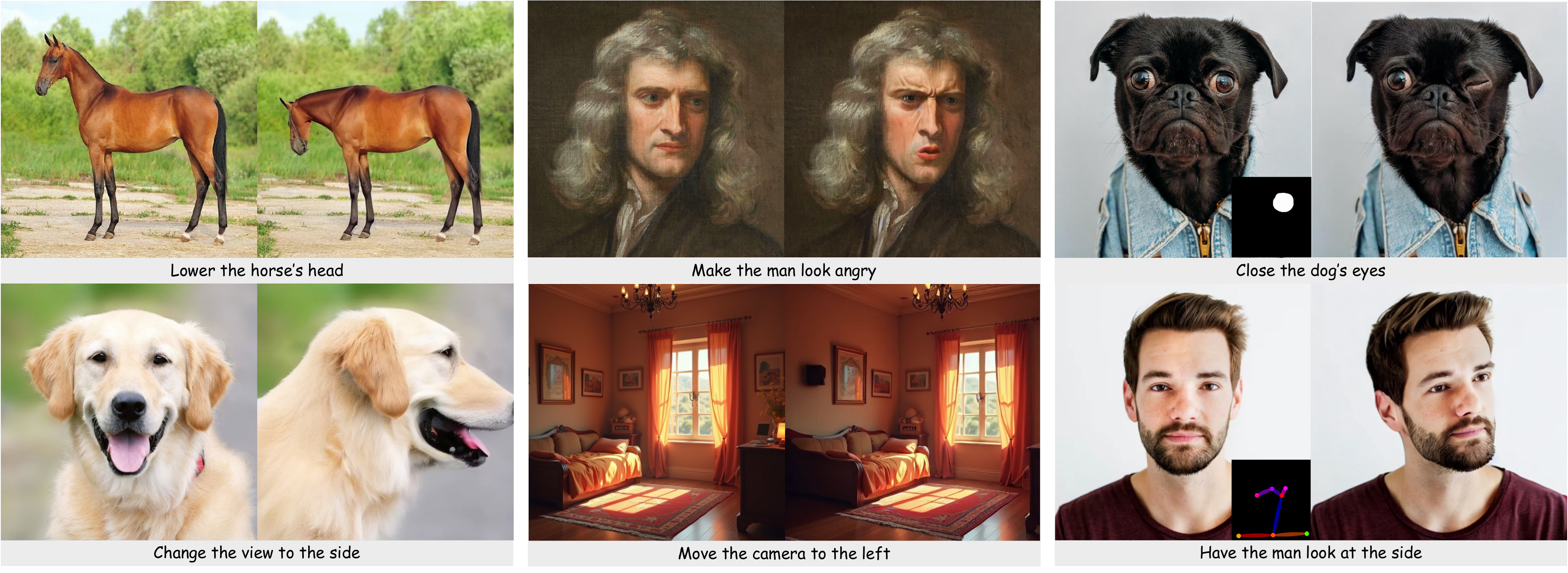}
    \captionof{figure}{
    We propose \textbf{InstructMove}, an instruction-based image editing model trained on frame pairs from videos with instructions generated by Multimodal LLMs. Our model excels at non-rigid editing, such as adjusting subject poses, expressions, and altering viewpoints, while maintaining content consistency. Additionally, our method supports precise, localized edits through the integration of masks, human poses, and other control mechanisms.
    \label{fig:teaser}
    }
\end{center}%
}]

\begin{abstract}
This paper introduces a novel dataset construction pipeline that samples pairs of frames from videos and uses multimodal large language models (MLLMs) to generate editing instructions for training instruction-based image manipulation models. Video frames inherently preserve the identity of subjects and scenes, ensuring consistent content preservation during editing. Additionally, video data captures diverse, natural dynamics—such as non-rigid subject motion and complex camera movements—that are difficult to model otherwise, making it an ideal source for scalable dataset construction. Using this approach, we create a new dataset to train \textbf{InstructMove}, a model capable of instruction-based complex manipulations that are difficult to achieve with synthetically generated datasets. Our model demonstrates state-of-the-art performance in tasks such as adjusting subject poses, rearranging elements, and altering camera perspectives. The project page is available \href{https://ljzycmd.github.io/projects/InstructMove/}{here}.

\end{abstract}    
\section{Introduction}
\label{sec:intro}
Recent years have seen remarkable progress in text-to-image (T2I) generation~\cite{dhariwal2021diffusion, nichol2021glide, rombach2022high, ramesh2022hierarchical, saharia2022photorealistic}, with state-of-the-art methods now capable of producing high-quality and hyper-realistic images. 
The key to this success lies in training on vast datasets containing billions of images from the internet, paired with text descriptions generated by image captioning models~\cite{radford2021learning}. 
In contrast, editing an existing image based on textual instructions remains challenging, with output quality still falling short and a limited range of editing types available.
A major bottleneck in this area is the difficulty of obtaining a comparably large dataset of source-target-instruction triplets for training.

\begin{figure}[!t]
    \centering
    \includegraphics[width=\linewidth]{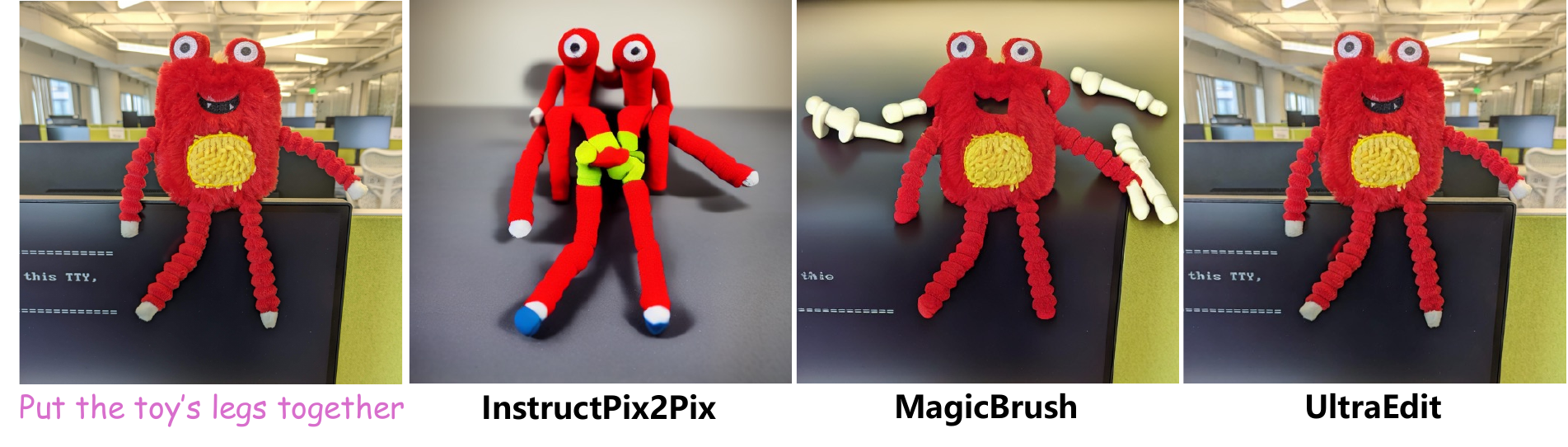}
    \caption{Existing methods struggle with complex edits on real photos, such as non-rigid transformations. They often either fail to follow the editing instructions or produce inconsistent outputs.}
    \label{fig:motivation}
\end{figure}

Previous approaches~\cite{kim2022diffusionclip, couairon2022diffedit, meng2021sdedit, hertz2022prompt, tumanyan2023plug, mokady2023null, cao2023masactrl} attempted to adapt pretrained T2I models without fine-tuning, using noise inversion to effectively ``regenerate'' the target image with modified textual prompts. However, these tuning-free methods are often slow and lack robustness. InstructPix2Pix~\cite{brooks2023instructpix2pix} introduced a method to bootstrap a dataset of source images into editing triplets by generating editing instructions using a language model and creating target images using tuning-free techniques. Subsequent works~\cite{zhang2024magicbrush, huang2024smartedit, sheynin2024emu, geng2024instructdiffusion} have aimed to refine this process or filter the generated outputs. 
Nonetheless, a fundamental limitation remains: target images are synthetically generated, which constrains the model's potential due to data quality issues. Consequently, existing methods often struggle to preserve the appearance of edited subjects or to perform precise, detailed edits, as shown in Fig.~\ref{fig:motivation}.

In this paper, we present a novel approach for creating large-scale, instruction-based image-editing datasets using real images sampled from internet videos, 
capturing naturally realistic transformations from source to target.
Our key insight is that video frames capture rich information on how things move — pose changes, elements shifts, and camera movements — making them an ideal data source for training models for image manipulation. We then leverage powerful multi-modal large language models (MLLMs), such as GPT-4o~\cite{achiam2023gpt} and LLaVA~\cite{liu2024visual} to generate descriptive text at or above human-level quality.

Our data construction starts with extracting video frames that undergo realistic yet complex transformations. Next, we use MLLMs to analyze these transformations and generate precise editing instructions. Additionally, we introduce a novel spatial conditioning strategy: instead of conventional channel concatenation, we concatenate the reference image with the noise input along the spatial dimension. This approach enables the model to access the reference image through cross-attentions across the network, greatly enhancing its ability to perform flexible edits while preserving the appearance of the source image.

By fine-tuning pretrained text-to-image models (\eg, Stable Diffusion~\cite{rombach2022high}) on our constructed dataset with spatial conditioning strategy, we enable natural-language-driven image manipulations that were previously challenging. These include adjusting subject poses, rearranging elements, or altering camera perspectives, all while maintaining the integrity and details of the original image (see Fig.~\ref{fig:teaser}). Our approach also seamlessly integrates with mask-based guidance for local editing, and also with other control types~\cite{zhang2023adding, mou2024t2i}, allowing for versatile modifications.

Our contributions can be summarized as follows:
\begin{itemize}
    \item We introduce a data construction pipeline that uses video frames and MLLMs to create source-target-instruction triplets for training image manipulation models.
    \item We introduce a spatial conditioning strategy that conditions the reference image alongside the noise map to improve content preservation while enabling flexible edits.
    \item Our approach enables a wider range of text-based image manipulations, including pose adjustments, element rearrangements, and perspective changes, while also supporting precise, localized edits through integration with additional control mechanisms.
\end{itemize}

\section{Related Work}
\label{sec:related_work}

\subsection{Zero-shot Image Editing with Diffusion Models}

Diffusion models~\cite{sohl2015deep, song2019generative, ho2020denoising} have become the leading approach in text-to-image generation due to their ability to effectively translate noise into structured data distributions. Leveraging these pretrained models, several image editing techniques have been developed by manipulating the sampling process~\cite{kim2022diffusionclip, meng2021sdedit, couairon2022diffedit, mokady2023null, kwon2022diffusion, parmar2023zero, huberman2024edit, brack2024ledits++}, modifying the internal architecture of the denoising network~\cite{hertz2022prompt, tumanyan2023plug, cao2023masactrl}, or incorporating additional optimization steps~\cite{kawar2023imagic, li2023layerdiffusion, zhang2023forgedit}.

For example, SDEdit adds noise to the image and then generates the desired edit by conditioning on a target prompt during the reverse denoising process. Other approaches, such as Prompt-to-Prompt~\cite{hertz2022prompt}, Plug-and-Play~\cite{tumanyan2023plug}, and MasaCtrl~\cite{cao2023masactrl}, manipulate the attention mechanism after inverting the image to noise using DDIM inversion~\cite{song2020denoising}, allowing for edits that align with the desired prompt. More recent works~\cite{mokady2023null, ju2023direct, han2024proxedit, deutch2024turboedit, wu2024turboedit} have refined the inversion process to enhance editing robustness and accuracy. While these zero-shot methods offer flexibility in editing images, they are often too slow and lack robustness, making them less suitable for practical applications.

\subsection{Training Instruction-based Editing Models}

Compared to modifying target image descriptions, a more intuitive and user-friendly editing method is to use direct instructions (\eg, “Have the man look at the camera”) to guide the editing process. 
InstructPix2Pix~\cite{brooks2023instructpix2pix} pioneered this concept by fine-tuning Stable Diffusion~\cite{rombach2022high} with editing instructions. To handle more complex instructions, methods like MGIE~\cite{fu2023guiding} and SmartEdit~\cite{huang2024smartedit} integrate Multimodal LLMs to improve the understanding of editing instructions.

A major challenge in training instruction-based editing models is constructing paired image data with corresponding instructions. InstructPix2Pix tackles this by using language models, such as GPT-3~\cite{brown2020language}, to generate both an edited image caption and an editing instruction. These captions are converted into images using Prompt-to-Prompt~\cite{hertz2022prompt}. MagicBrush~\cite{zhang2024magicbrush} constructs a small dataset by performing edits using DALL$\cdot$E 2~\cite{ramesh2022hierarchical}. InstructDiffusion~\cite{geng2024instructdiffusion} augments datasets with synthetic pairs for tasks like object removal and replacement. EmuEdit~\cite{sheynin2024emu} improves data pair construction using LLama 2~\cite{touvron2023llama} and local masks with Prompt-to-Prompt, combining these with other vision datasets to build a large-scale dataset.

However, in existing large-scale editing datasets, most target images are generated using synthetic methods, which often introduces artifacts and significant appearance deviations from the source images. These datasets are typically limited to high-level edits such as object overlay and style transfer. As a result, models trained on these datasets struggle with complex edits on real photos, such as non-rigid transformations and viewpoint changes. In contrast, our approach leverages real video frames as source and target images, using Multimodal LLMs to generate instructions directly from these frames. This allows for the construction of a more realistic dataset, capturing complex transformations and overcoming the limitations of previous methods.

\section{Dataset Construction}
\label{sec:dataset}

\begin{figure}
    \centering
    \includegraphics[width=\linewidth]{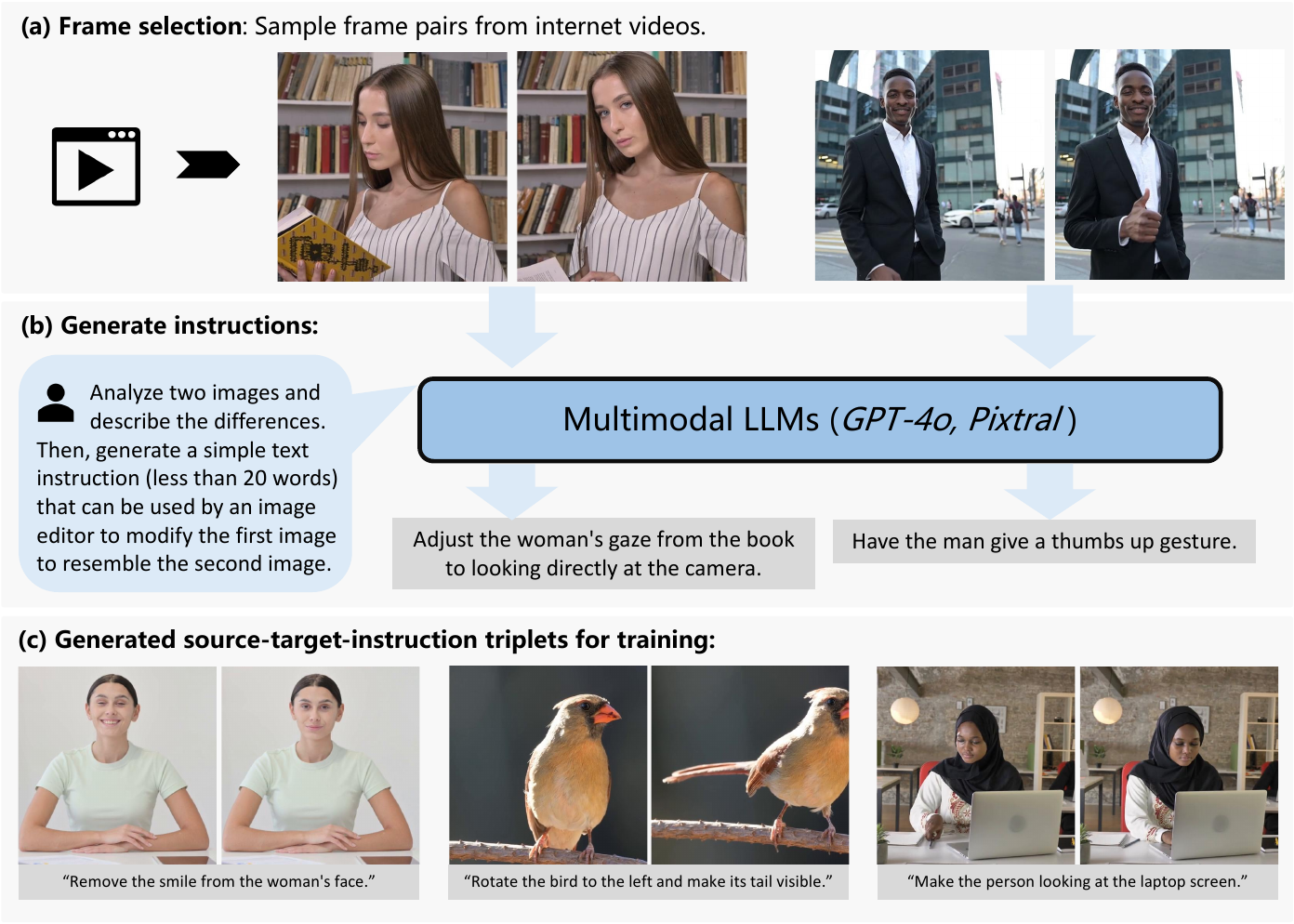}
    \vspace{-2em}
    \caption{\textbf{Our data construction pipeline.} (a) We begin by sampling suitable frame pairs from videos, ensuring realistic and moderate transformations. (b) These frame pairs are used to prompt Multimodal Large Language Models (MLLMs) to generate detailed editing instructions. (c) This process results in a large-scale dataset with realistic image pairs and precise editing instructions.
    }
    \label{fig:data_pipeline}
    \vspace{-2mm}
\end{figure}

Unlike existing instruction-based editing datasets, which rely on tuning-free methods to generate target images, we propose a novel data construction pipeline that leverages real video frame pairs. By extracting frame pairs that capture meaningful transformations and using Multimodal Large Language Models (MLLMs) to generate detailed editing instructions, we create a more realistic and high-quality editing dataset. In Sec.~\ref{subsec:filter}, we outline our approach for extracting suitable frame pairs from videos, followed by Sec.~\ref{subsec:prompt}, where we describe how we use MLLMs to generate precise editing instructions. Finally, Sec.~\ref{subsec:analysis} provides an in-depth analysis of the resulting dataset. An overview of the data construction pipeline is illustrated in Fig.~\ref{fig:data_pipeline}.

\subsection{Sampling Frame Pairs from Videos}
\label{subsec:filter}
In a video, the same subjects and elements can appear in different ways, move naturally, or be shown from different angles. For example, a person might change their posture, expression, or movement over time. These changes provide useful clues about how a scene or perspective might look with small adjustments, making videos a great source for picking frame pairs to use in image editing tasks. However, it’s important to sample the frame pairs with proper criteria to make sure they are suitable for training.
\vspace{2mm}

\noindent\textbf{Initial sampling of paired frames.} We begin by captioning each video to generate an overall description. Videos containing specific keywords (\eg, landscape, abstract, still) that are not suitable for image editing are filtered out. We then sample two frames $(I^s, I^e)$ with a fixed time interval (\ie, 3 seconds) to ensure sufficient transformations occur in the objects of interest.

\vspace{2mm} \noindent\textbf{Motion-based filtering.} To refine the selection, we apply motion filtering to exclude frames with either too little or too much movement. Specifically, we use RAFT~\cite{teed2020raft} to compute the optical flow $O_{I^s\rightarrow I^e}$ from $I^s$ to $I^e$ and derive the flow magnitude $M=\sqrt{O^2_x + O^2_y}$. Frame pairs with moderate movement are retained, while those with excessive or minimal motion are discarded. To further ensure consistency, we assess background changes by computing a background occlusion mask through backward warping of $I^e$ to $I^s$ using the estimated flow. By measuring the occlusion mask ratio, we filter out frames with significant background changes (\ie, large masked regions). This motion filtering pipeline allows us to obtain image pairs with realistic, moderate transformations.

\subsection{Instruction Generation with MLLMs}
\label{subsec:prompt}
Next, we describe how we generate accurate editing instructions $C$ given frame pairs $(I^s, I^e)$. Previous approaches~\cite{brooks2023instructpix2pix, sheynin2024emu} rely solely on text-based methods, using Large Language Models (LLMs) to generate instructions and captions of target images purely from the source caption. This approach significantly limits the variety of edits that can be captured in the dataset. 
As a result, these methods are often constrained to high-level edits, such as adding, removing, or replacing elements, or changing the style.

To address this limitation, we utilize \textbf{multimodal} LLMs, such as GPT-4o~\cite{achiam2023gpt} or Pixtral-12B~\cite{agrawal2024pixtral}, to directly analyze frame pairs and generate editing instructions. Specifically, MLLMs are prompted to examine the differences between the source and target images, focusing on changes in subjects, relative positions, camera angles, and background. Based on these observations, the models generate detailed, context-specific instructions. 

To improve clarity, we instruct the MLLMs to use absolute terms (\eg, “Move the bee to the center of the flower” instead of “Move the bee to the position in the target image”) and to begin instructions with an action verb (\eg, Change, Move, Adjust) for better usability in real-world applications. Additionally, to ensure high-quality instructions, we allow the MLLMs to reject frame pairs where the edits are too complex to describe accurately, and we exclude such pairs from our dataset.

\begin{table}[!t]
    \centering
    \small
    \setlength{\tabcolsep}{2.5pt}
    \resizebox{\linewidth}{!}{
    \begin{tabular}{lcccccc}
    \toprule
    \multirow{2}{*}{Datasets} & Real & \multirow{2}{*}{\# Edits} & \multicolumn{4}{c}{Editing Types} \\
    \cmidrule{4-7}
    & Target? & & Non-rigid & Viewpoint & Add/Remove & Style \\
    \midrule
    InstructPix2Pix~\cite{brooks2023instructpix2pix} & $\newcrossmark$ & 0.3M & $\newcrossmark$  & $\newcrossmark$ & $\newcheckmark$ & $\newcheckmark$\\
    MagicBrush~\cite{zhang2024magicbrush} & $\newcrossmark$ & 10K & $\newcrossmark$  & $\newcrossmark$ & $\newcheckmark$ & $\newcheckmark$\\
    EmuEdit~\cite{sheynin2024emu} & $\newcrossmark$ & 10M & $\newcrossmark$  & $\newcrossmark$ & $\newcheckmark$ & $\newcheckmark$\\
    UltraEdit~\cite{zhao2024ultraedit} & $\newcrossmark$ & 4.1M & $\newcrossmark$  & $\newcrossmark$ & $\newcheckmark$ & $\newcheckmark$\\
    \midrule
    Ours & $\newcheckmark$ & 6M & $\newcheckmark$  & $\newcheckmark$ & $\newcrossmark$ & $\newcrossmark$\\
    \bottomrule 
    \end{tabular}
    }
    \caption{\textbf{Comparison with existing image-editing datasets.} Our dataset is the first large-scale dataset with real target images while supporting complex editing types, such as non-rigid transformations and camera viewpoint adjustments.}
    \vspace{-4mm}
    \label{tab:editing_types}
\end{table}

\subsection{Dataset Analysis}
\label{subsec:analysis}
In total, after filtering, we sampled 6 million image pairs from videos, each with appropriate levels of motion and transformation. All frames are captioned using Pixtral-12B~\cite{agrawal2024pixtral}. The edits captured in our dataset focus on realistic transformations, such as non-rigid changes of subjects/objects, rearranging elements, and altering camera perspectives—types of edits that are largely absent in existing datasets. Examples can be seen in Fig.~\ref{fig:data_pipeline}, with comparisons to prior datasets presented in Table~\ref{tab:editing_types}. 

Our dataset can be combined with existing ones to train models that handle both realistic edits introduced in this paper, and the artistic edits explored in previous works (\eg, converting a photo into a painting), as shown in the supplementary material.

As illustrated in Fig.~\ref{fig:data_pipeline}, some video frames may include subtle changes that are not fully captured in the generated instructions. However, our experiments show that these small discrepancies do not significantly impact the training process, as the model focuses on learning the major transformations between frame pairs.
\section{Method}
\label{sec:method}

\begin{figure}
    \centering
    \includegraphics[width=\linewidth]{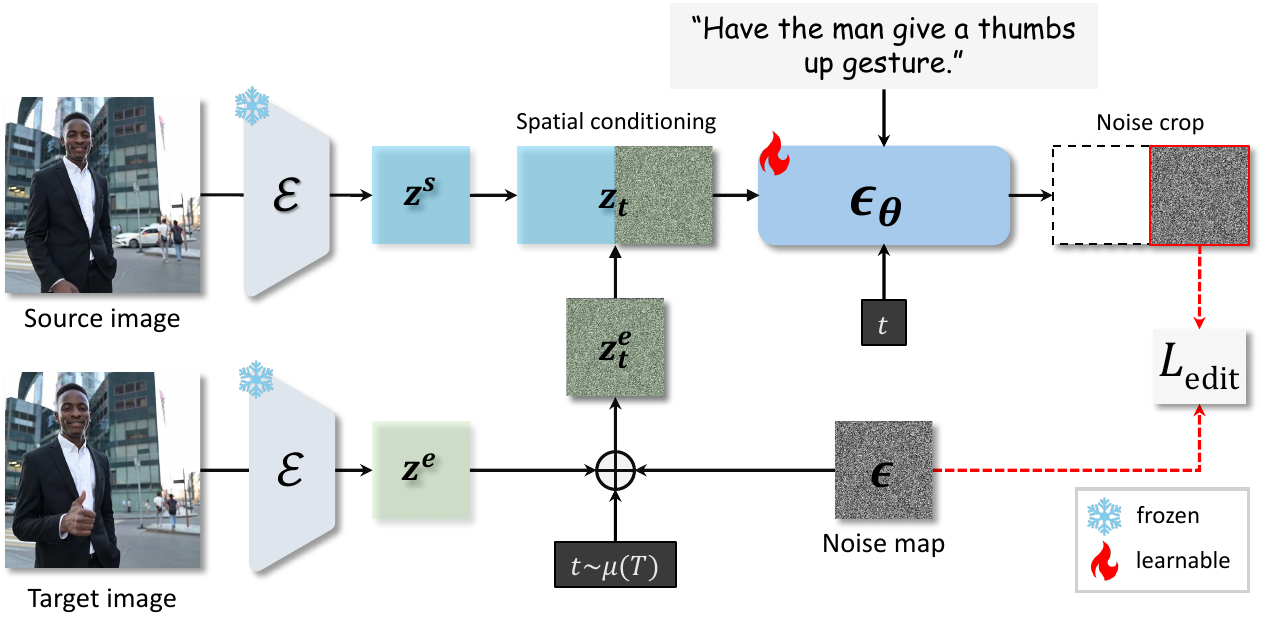}
    \vspace{-2em}
    \caption{\textbf{Overview of the proposed model architecture for instruction-based image editing.} The source and target images are first encoded into latent representations $z^s$ and $z^e$ using a pretrained encoder. The target latent $z^e$ is then transformed into a noisy latent $z^e_t$ through the forward diffusion process. We concatenate the source image latent and the noisy target latent along the width dimension to form the model input, which is fed into the denoising U-Net $\epsilon_\theta$ to predict a noise map. The right half of the output, corresponding to the noisy target input, is cropped and compared with the original noise map. 
    }
    \label{fig:main_arch}
    \vspace{-4mm}
\end{figure}

Using the constructed dataset, we fine-tune a pretrained T2I model for instruction-based image manipulation, enabling edits such as non-rigid transformations of subjects, object repositioning, and changes in viewpoint. Unlike previous methods, in Sec.~\ref{subsec:condition} we introduce a novel spatial conditioning approach that conditions the pretrained T2I models on reference images without requiring any architectural modifications. This technique allows instructions to control the target structure while preserving the identity and content of the original image. Since our method does not alter the underlying architecture of the pretrained T2I models, it seamlessly integrates with additional control mechanisms, such as masks or ControlNets~\cite{zhang2023adding}, to enable even more precise and localized edits (Sec.~\ref{subsec:mask}).

\subsection{Finetuning T2I Models for Image Manipulation}
\label{subsec:condition}
Existing instruction-based editing models~\cite{brooks2023instructpix2pix, sheynin2024emu, huang2024smartedit, fu2023guiding} 
typically condition T2I models on a reference image by concatenating it with the noise map along the \textit{channel dimension} as input to the denoising network. 
However, this channel-based conditioning method spatially aligns the reference and target images, which limits the model's flexibility in making structural changes, such as non-rigid transformations or viewpoint adjustments.

To address this, we introduce \textbf{Spatial Conditioning}, which conditions the model by concatenating the source image with the noise map along the \textit{spatial dimension}.

Given a triplet $(I^s, I^e, C)$, we first encode both images into latent representations $z^s$ and $z^e$ using a pretrained variational autoencoder~\cite{kingma2013auto}. The target latent $z^e$ is then transformed into a noisy latent $z^e_t$ via the forward diffusion process. We concatenate the source image latent with the noisy target latent along the width dimension to create the model input $z_t = \text{Concat}_{width}([z^s, z^e_t])$. This input, which doubles the width of the noisy latent, is fed into the denoising U-Net $\epsilon_\theta$ to predict a noise map of the same width. We then crop the right half of the output corresponding to the noisy input and compute the denoising loss:
\begin{equation} \label{eq}
\vspace{-2mm}
\mathcal{L}_{\text{Edit}} = \mathbb{E}_{z_t, C, t, \epsilon} \left[\|\epsilon - \text{Crop}_{width}(\epsilon_\theta(z_t, C, t))\|^2\right].
\end{equation}

This spatial conditioning approach, while seemingly similar to channel conditioning, offers critical advantages for image editing tasks. First, it does not spatially align the reference image with the noisy input, allowing the denoising network to process the condition and the noisy input in parallel. This enables the network to attend to any region of the condition image at each cross-attention layer, facilitating structural changes in the target image. Moreover, since the network has access to the features of the condition image throughout all layers, it can better preserve the details and identity of the source image, leading to more accurate and realistic edits.

\subsection{Incorporating Additional Controls}
\label{subsec:mask}
Once trained, our editing model can perform a wide range of edits based on user-provided instructions. However, text instructions alone are often insufficiently precise, making additional controls essential for accurate image editing.

\vspace{2mm}
\noindent\textbf{Integration with masks.} Similar to zero-shot inpainting in T2I models, our approach supports mask-based localization to edit specific regions of an image. During inference, we blend the updated latent with the reference latent using a mask to control the area of modification:
\begin{equation}
    \label{eq:masked_denoising}
    \vspace{-2mm}
    z^*_{t-1} = (1 - m) \cdot z^s_{t-1} + m \cdot z_{t-1},
\end{equation}
where $z^s_{t-1}$ is the noisy latent from the forward diffusion of the source image latent $z^s$, and $m$ is the mask resized to match the latent resolution.

\noindent\textbf{Integration with other spatial controls.} For more complex edits, additional visual cues beyond text can greatly enhance control. Users can provide sketches, or adjust key points of a skeleton to modify poses, for example. To enable this, we integrate controllable diffusion models like ControlNet~\cite{zhang2023adding} and T2I-Adapter~\cite{mou2024t2i}. Crucially, because our model retains the architecture of the original T2I model, we can directly use these control techniques with our trained image-editing models, incorporating spatial maps for finer control. This compatibility is not possible with previous instruction-based editing models like InstructPix2Pix.
\section{Experiments}
\label{sec:exp}
In Sec.~\ref{subsec:setup}, we outline the experimental setup, including the evaluation benchmark and quantitative metrics. In Sec.~\ref{subsec:comparison}, we compare with state-of-the-art text-guided image editing methods both quantitatively and qualitatively. Sec.~\ref{subsec:control} demonstrates our model's ability to incorporate additional controls for more precise editing. Finally, we assess the impact of our dataset and conditioning strategy in Sec.~\ref{subsec:ablation}.

\subsection{Experimental Setup}
\label{subsec:setup}

\vspace{2mm}
\noindent\textbf{Implementation Details.}
We fine-tune Stable Diffusion V1.5~\cite{rombach2022high} with the proposed spatial conditioning strategy on our newly constructed dataset to enable various types of complex image edits. Input images have a resolution of $512\times 512$. The model is trained for 100,000 iterations with a constant learning rate of $1\times 10^{-4}$, using the Adam optimizer to update model parameters. Training is conducted on 8 NVIDIA A100 GPUs with a total batch size of 256. During sampling, we employ the DDIM~\cite{song2020denoising} scheduler with 50 steps to generate the final edited images. Note that mask and additional spatial controls are optional; when unspecified, results are generated solely based on textual instructions. 

\vspace{2mm}
\noindent\textbf{Evaluation benchmark.}
Existing instruction-based image editing benchmarks, such as MagicBrush~\cite{zhang2024magicbrush} and Emu-Edit~\cite{sheynin2024emu}, primarily focus on editing tasks that maintain the original image structure, such as style modification, object replacement, and color or texture changes. These benchmarks are therefore unsuitable for evaluating non-rigid, consistent image editing tasks that alter the image structure while preserving identity and appearance details. To address this gap, we created a specialized test set tailored for the non-rigid editing tasks presented. Our benchmark consists of 50 images, each paired with human-curated editing instructions that specify modifications in pose, expression, viewpoint, shape, position, and other structural transformations. The benchmark excludes any local or global style or appearance edits. Examples of input images and their corresponding instructions are shown in Fig.~\ref{fig:qualitative_results}. More details are in the supplementary material.

\vspace{2mm}
\noindent\textbf{Evaluation metrics.}
For evaluation, we employ metrics to assess two main aspects of image editing: alignment with text instructions and fidelity to the source image. To evaluate instruction adherence, following previous works, we first generate captions for the target image using a large language model (LLM) based on the source image’s caption and the given instructions. We then 
assess the agreement between caption changes and actual image changes (CLIP-D). However, this metric relies on generating captions for target images, which may not directly measure alignment with the original instructions.
To address this, we propose a new metric: we prompt MLLMs to generate instructions based on the input and output images for each method, then compute the CLIP distance between these generated instructions and the original ones, denoted as CLIP-Inst. 
For evaluating faithfulness, we calculate the CLIP feature distance between the input and output images (CLIP-I).

We also collect human feedback from 40 participants to evaluate the quality of the edited images. Each participant is shown a subset of 20 examples, including the outputs from each method, alongside the source image and the editing instruction. Participants are asked to choose the image that “best follows the editing instruction while maintaining the highest quality and preserving the original content.”

\subsection{Baseline Comparisons}
\label{subsec:comparison}
\vspace{2mm}
\noindent\textbf{Methods for comparison.}
We compare our proposed method against state-of-the-art text-guided image editing approaches with publicly available models, including zero-shot description-based methods: NullTextInversion~\cite{mokady2023null}, MasaCtrl~\cite{cao2023masactrl}, and Imagic~\cite{kawar2023imagic}, as well as instruction-based methods: InstructPix2Pix~\cite{brooks2023instructpix2pix}, MagicBrush~\cite{zhang2024magicbrush}, and UltraEdit~\cite{zhao2024ultraedit}. For description-guided methods, we generate the target description by captioning the input image and deriving the description from the instructions using an LLM.

\begin{table}[!t]
    \centering
    \small
    \setlength{\tabcolsep}{2.5pt}
    \resizebox{\linewidth}{!}{
    \begin{tabular}{lccc}
    \toprule
    \textbf{Method} & \textbf{CLIP-D} $\uparrow$ & \textbf{CLIP-Inst}$\uparrow$ & \textbf{CLIP-I}$\uparrow$ \\
    \midrule
    NullTextInversion~\cite{mokady2023null} & 0.0660 & 0.7648 & 0.9063 \\
    MasaCtrl~\cite{cao2023masactrl} & 0.0436 & 0.8527 & 0.9160 \\
    Imagic~\cite{kawar2023imagic} & 0.0508 & 0.8645 & 0.8379 \\
    InstructPix2Pix~\cite{brooks2023instructpix2pix} & 0.0887 & 0.8569 & \textbf{0.9380} \\
    MagicBrush~\cite{zhang2024magicbrush} & 0.0972 & 0.8648 & 0.9318 \\
    UltraEdit~\cite{zhao2024ultraedit} & 0.0824 & 0.8571 & 0.9184 \\
    \midrule
    Ours & \textbf{0.1361} & \textbf{0.8724} & 0.9275 \\
    \bottomrule
    \end{tabular}
    }
    \caption{Quantitative comparison with state-of-the-art text-guided image editing methods on our collected benchmark.}
    \label{tab:quantitative_results}
\end{table}

\begin{table}[!t]
    \centering
    \tabcolsep=3.5pt
    \resizebox{\linewidth}{!}{
    \begin{tabular}{lcccc}
    \toprule
     & Imagic~\cite{kawar2023imagic} & InstructPix2Pix~\cite{brooks2023instructpix2pix} & MagicBrush~\cite{zhang2024magicbrush} & Ours \\
    \midrule
    \textbf{Preference} & 5.0\% & 3.25\% & 4.13\% & \textbf{87.62\%} \\
    \bottomrule
    \end{tabular}
    }
    \caption{\textbf{Human preference results.} Percentage indicates how often each method was selected as the best result.}
    \label{tab:user}
\end{table}

\begin{figure*}
    \centering
    \includegraphics[width=\linewidth]{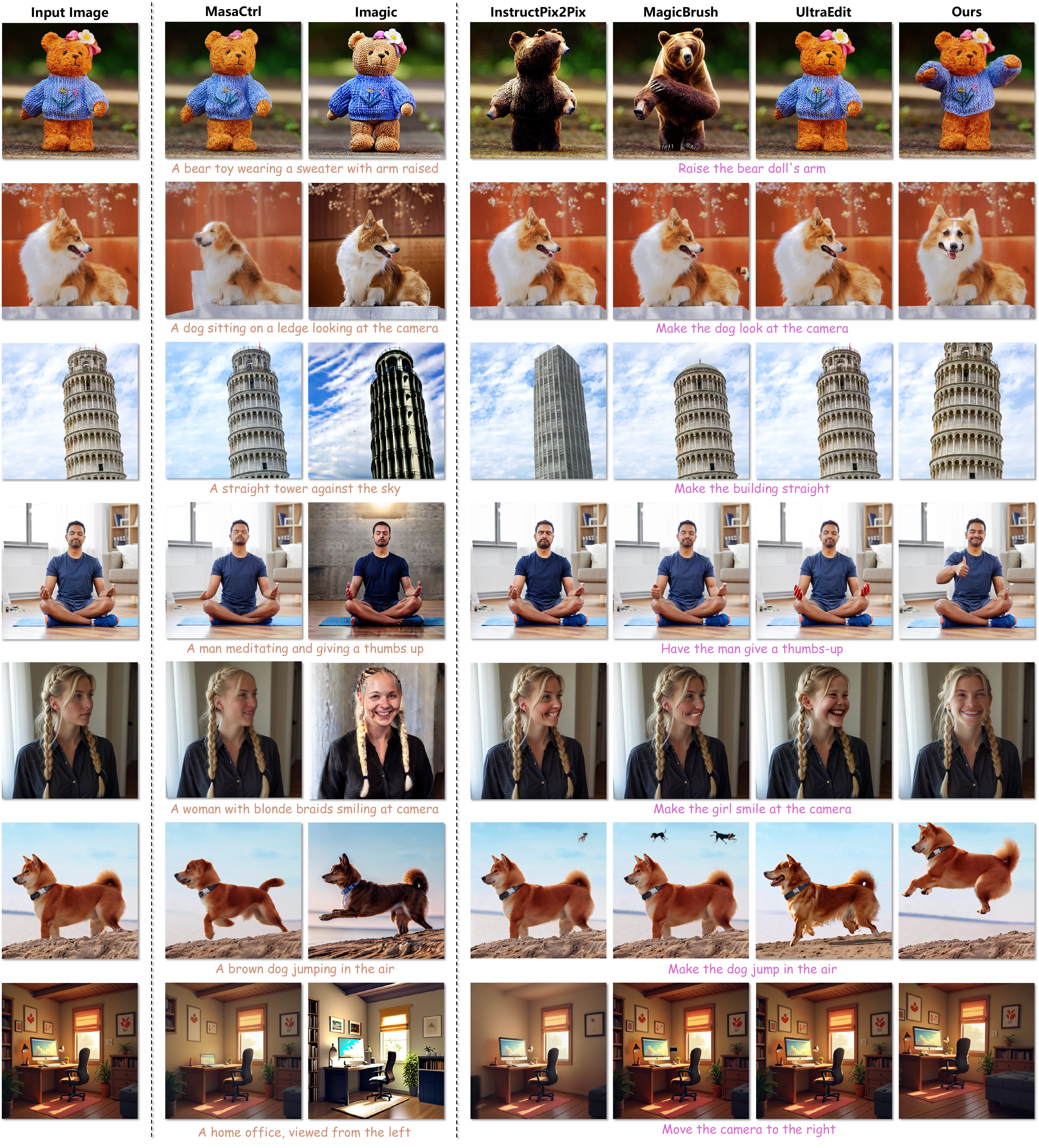}
    \caption{\textbf{Qualitative comparison with state-of-the-art image editing methods}, including both description-based and instruction-based approaches. Existing methods struggle with complex edits such as non-rigid transformations (\eg, changes in pose and expression), object repositioning, or viewpoint adjustments. They often either fail to follow the editing instructions or produce images with inconsistencies, such as identity shifts. In contrast, our method, trained on real video frames with naturalistic transformations, successfully handles these edits while maintaining consistency with the original input images.}
    \label{fig:qualitative_results}
\end{figure*}

\vspace{2mm}
\noindent\textbf{Quantitative results.}
\Cref{tab:quantitative_results} presents the quantitative evaluation of both description- and instruction-based methods on our editing benchmark. A method that perfectly preserves content but fails to follow the editing instructions is still inadequate. For instance, InstructPix2Pix and MagicBrush show higher CLIP-I scores than our approach because they struggle with non-rigid edits and viewpoint changes. As a result, they often leave the input image mostly unchanged (see~\cref{fig:qualitative_results}), leading to high content preservation scores. In contrast, our method achieves the highest scores in instruction alignment while still maintaining strong content preservation. This is further validated by the user study in Table~\ref{tab:user}, where in over 87.62\% of cases, users preferred our method over the other four approaches.

\vspace{2mm}
\noindent\textbf{Qualitative results.}
We present a qualitative comparison of various methods in~\cref{fig:qualitative_results}. Since prior approaches are primarily designed for editing tasks that preserve the original image structure, they struggle with non-rigid edits (such as changes in pose and expression), object repositioning, or viewpoint adjustments. These methods often fail to follow such editing instructions, either producing images identical to the source or generating edits that do not align with the specified instructions. Even in cases where the instructions are somewhat followed, they frequently fail to maintain content consistency.
In contrast, our method, trained on real video frames with naturalistic transformations, is capable of handling these complex edits while preserving consistency with the input images.

\begin{figure*}
    \centering
    \includegraphics[width=\linewidth]{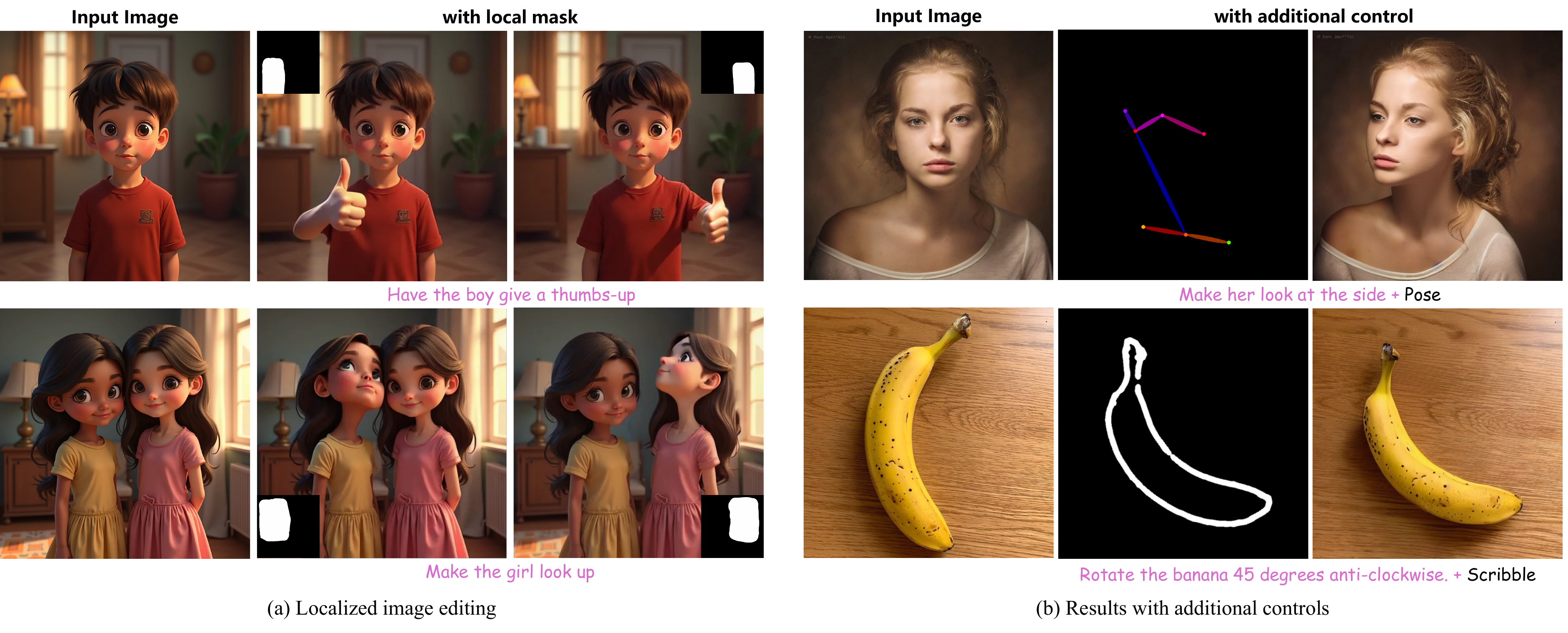}
    \caption{\textbf{Qualitative results of our method with additional controls.} (a) Our model can utilize a mask to specify which part of the image to edit, enabling localized adjustments and resolving ambiguities in the instructions. (b) When combined with ControlNet, our model can accept additional inputs, such as human poses or rough sketches, to achieve precise edits in subject poses or object positioning. This level of control is not possible with previous methods.}
    \label{fig:results_mask_control}
\end{figure*}

\subsection{Results with Additional Controls}
\label{subsec:control}
We demonstrate the incorporation of additional controls for precise editing in Fig.~\ref{fig:results_mask_control}. Instructions alone can sometimes be ambiguous regarding which part of the image to edit; in such cases, a mask can be used for localized adjustments (Fig.~\ref{fig:results_mask_control} (a)). In other scenarios, instructions alone may not suffice, especially when precise control is needed for adjustments like altering a subject’s pose. Users can address this by directly adjusting control points to define a skeleton pose or providing a rough sketch of the desired composition. We show that our approach can seamlessly integrate ControlNet to utilize such controls for precise editing (Fig.~\ref{fig:results_mask_control} (b)).

\subsection{Ablations}
\label{subsec:ablation}

\begin{table}[!t]
    \centering
    \tabcolsep=5pt
    \begin{tabular}{lccc}
    \toprule
    \textbf{Method} & \textbf{CLIP-D}$\uparrow$ & \textbf{CLIP-Inst} $\uparrow$ & \textbf{CLIP-I} $\uparrow$ \\
    \midrule
    SC + IP2P data & 0.1277 & 0.8414 & 0.9094 \\
    CC + Our data  & 0.0853 & 0.8679 & 0.8552 \\
    \midrule
    SC + Our data & \textbf{0.1361} & \textbf{0.8724} & \textbf{0.9275}  \\
    \bottomrule
    \end{tabular}
    \caption{
    \textbf{Ablation on the training dataset and conditioning approach.} Training on the InstructPix2Pix dataset~\cite{brooks2023instructpix2pix} leads to poorer instruction alignment and content preservation. Additionally, our Spatial Conditioning (SC) method outperforms the standard Channel Conditioning (CC) approach.
    }
    \vspace{-2mm}
    \label{tab:ablation}
\end{table}

\begin{figure}
    \centering
    \includegraphics[width=\linewidth]{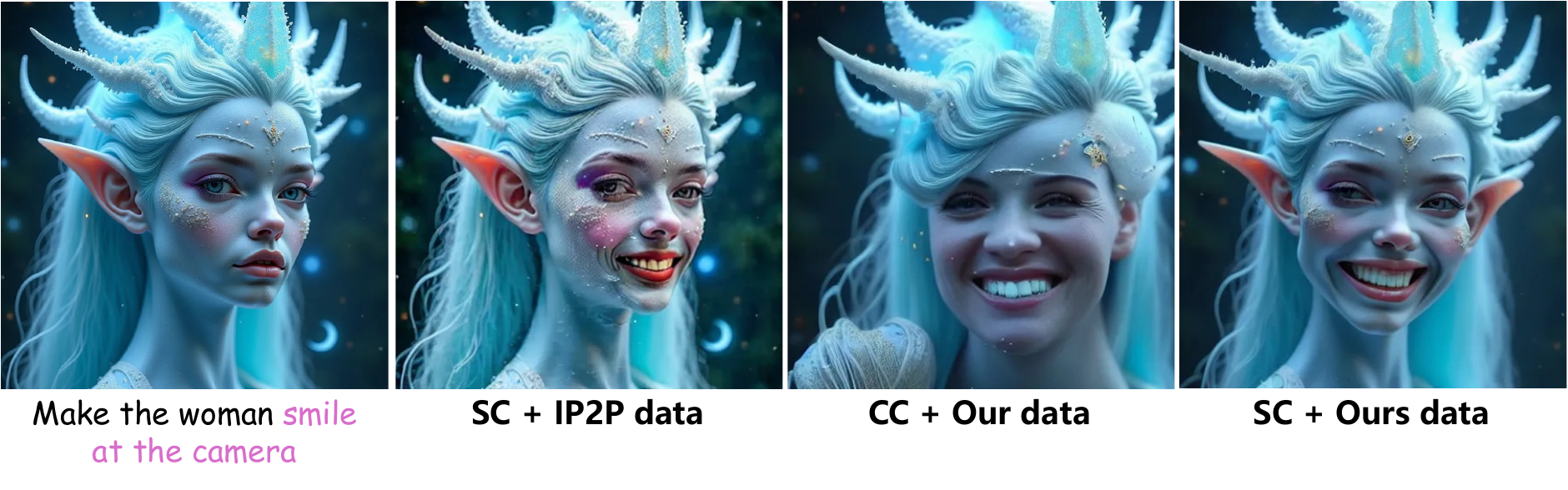}
    \vspace{-1em}
    \caption{\textbf{Qualitative ablation on the training dataset and conditioning approach.} Compared to models trained on the InstructPix2Pix dataset (IP2P)~\cite{brooks2023instructpix2pix} or using the standard Channel Conditioning (CC) method, our model trained on our dataset with the Spatial Conditioning (SC) approach demonstrates superior instruction alignment and content preservation.}
    \label{fig:ablation}
    \vspace{-4mm}
\end{figure}

We argue that our proposed video + MLLM dataset construction pipeline, along with the Spatial Conditioning (SC) approach, enables our model to perform complex image edits while maintaining appearance consistency. We evaluate the impact of each component in Table~\ref{tab:ablation} and Figure~\ref{fig:ablation}. 

\vspace{2mm}
\noindent\textbf{Dataset.} To demonstrate the importance of using real video frame pairs for training image editing models, we fine-tune a model with the same architecture as ours on the dataset proposed in InstructPix2Pix~\cite{brooks2023instructpix2pix}, where target images are synthetically generated using Prompt-to-Prompt~\cite{hertz2022prompt}. Compared to our model trained on InstructPix2Pix’s dataset (denoted as “SC + IP2P data”) , our method not only better follows editing instructions but also preserves the original image content more effectively.

\vspace{2mm}
\noindent\textbf{Spatial Conditioning.} We also fine-tune our model using our dataset, but with the Channel Conditioning (CC) method, where the reference latent is concatenated with the target latent along the channel dimension rather than the spatial dimension. As shown in Table~\ref{tab:ablation} and Figure~\ref{fig:ablation}, the proposed Spatial Conditioning approach significantly improves the model’s ability to apply complex edits while achieving higher quality and better content preservation.

\begin{figure}
    \centering
    \includegraphics[width=\linewidth]{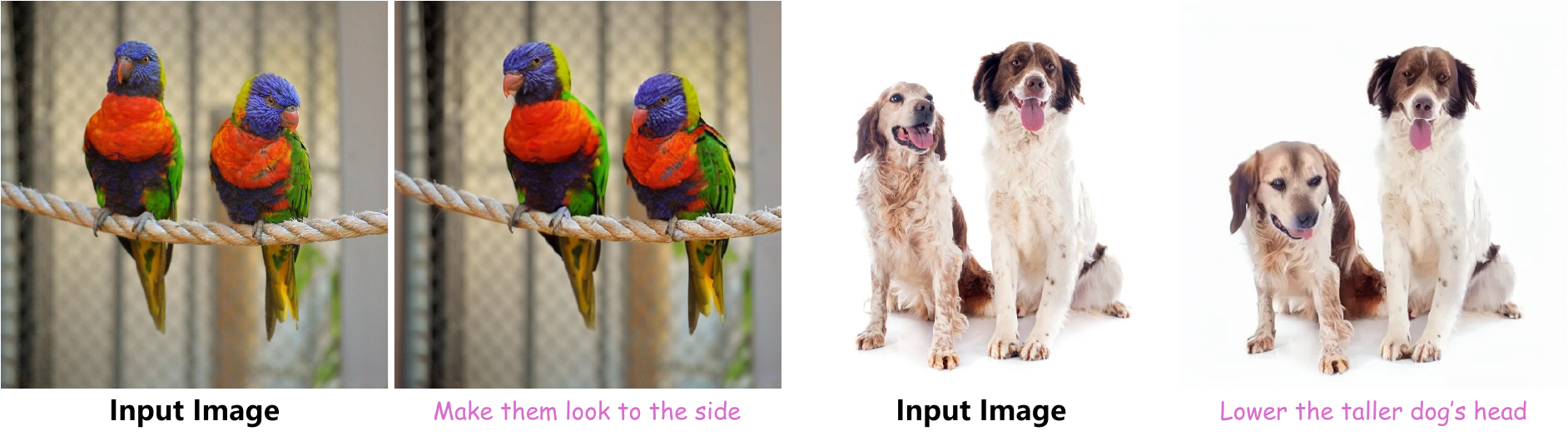}
    \vspace{-2em}
    \caption{\textbf{Limitations.} Left: Occasional unintended slight viewpoint changes. Right: Inability to accurately isolate specific objects based on instructions.}
    \label{fig:limitations}
    \vspace{-4mm}
\end{figure}

\section{Limitation}
Our approach introduces a novel data construction pipeline, but it still has some limitations. The quality of the dataset is influenced by the frame filtering process and the capabilities of the MLLMs. Occasionally, MLLMs may generate inaccurate instructions or fail to capture all transformations between frames, leading our editing model to apply unintended changes, such as slight viewpoint shifts that deviate from the original instructions. The effectiveness of our model also depends on the accuracy of training instructions and the comprehension abilities of pretrained T2I models, which can affect its performance on complex edits. Examples of these limitations are shown in Fig.~\ref{fig:limitations}. Moreover, since the transformations captured from video frames are restricted to realistic changes, our model cannot handle artistic edits like style transfer or object replacement. This limitation can be mitigated by integrating our dataset with others, as demonstrated in the supplementary material.

\section{Conclusion}
We present an approach for sampling video frames and utilizing MLLMs to generate editing instructions for training instruction-based image manipulation models. Unlike existing datasets, which rely on synthetically generated target images, our method leverages supervision signals from videos and MLLMs to support complex edits, such as non-rigid transformations and viewpoint changes, while preserving content consistency. Future work could focus on refining filtering techniques, either by improving MLLMs or incorporating human-in-the-loop processes, as well as integrating video data with other datasets to further enhance image editing capabilities.

{
    \small
    \bibliographystyle{ieeenat_fullname}
    \bibliography{main}
}


\end{document}